\documentclass[a4paper]{article}
\usepackage{INTERSPEECH2015}
\usepackage{epsfig}
\usepackage{algorithm,algorithmic}
\usepackage{amsmath,graphicx,paralist}
\usepackage{tabularx}
\usepackage{array}

\def\thline{\noalign{\hrule height 1pt}}
\newcolumntype{L}[1]{>{\raggedright\let\newline\\\arraybackslash\hspace{0pt}}m{#1}}
\newcolumntype{C}[1]{>{\centering\let\newline\\\arraybackslash\hspace{0pt}}m{#1}}
\newcolumntype{R}[1]{>{\raggedleft\let\newline\\\arraybackslash\hspace{0pt}}m{#1}}

\title{English Conversational Telephone Speech Recognition by Humans and Machines}
\twoauthors{\begin{tabular}{c}George Saon, Gakuto Kurata$^*$, Tom Sercu\\
Kartik Audhkhasi, Samuel Thomas, Dimitrios Dimitriadis\\
Xiaodong Cui, Bhuvana Ramabhadran, Michael Picheny\end{tabular}}{IBM Watson, Yorktown Heights, USA\\
$^*$IBM Research - Tokyo, Japan}{Lynn-Li Lim, Bergul Roomi, Phil Hall}{Appen, Sydney, Australia}

\ninept
\copyrightnotice{\copyright\ IEEE 2004}
\toappear{Submitted to Interspeech 2017, please do not redistribute}

\begin{document}
\maketitle

\begin{abstract} 

  One of the most difficult speech recognition tasks is accurate
  recognition of human to human communication. Advances in deep
  learning over the last few years have produced major speech
  recognition improvements on the representative Switchboard
  conversational corpus. Word error rates that just a few years ago
  were 14\% have dropped to 8.0\%, then 6.6\% and most recently
  5.8\%, and are now believed to be within striking range of human
  performance. This then raises two issues - what IS human
  performance, and how far down can we still drive speech recognition
  error rates? A recent paper by Microsoft suggests that we have
  already achieved human performance.  In trying to verify this
  statement, we performed an independent set of human performance
  measurements on two conversational tasks and found that human
  performance may be considerably better than what was earlier
  reported, giving the community a significantly harder goal to
  achieve. We also report on our own efforts in this area, presenting
  a set of acoustic and language modeling techniques that lowered the
  word error rate of our own English conversational telephone LVCSR
  system to the level of 5.5\%/10.3\% on the Switchboard/CallHome
  subsets of the Hub5 2000 evaluation, which - at least at the writing
  of this paper - is a new performance milestone (albeit not at what
  we measure to be human performance!). On the acoustic side, we use a
  score fusion of three models: one LSTM with multiple feature
  inputs, a second LSTM trained with speaker-adversarial multi-task
  learning and a third residual net (ResNet) with 25 convolutional layers
  and time-dilated convolutions.
  On the language modeling side, we use word and character LSTMs and
  convolutional WaveNet-style language models.

\end{abstract}


\section{Introduction}

With the performance of ASR systems inching ever closer to that of
humans, it is important to benchmark human performance accurately.
In~\cite{xiong16}, the authors claim a human word error rate (WER) of
5.9\%/11.3\% on the Switchboard/CallHome subsets (SWB/CH) of the NIST
Hub5 2000 evaluation testset. When compared with~\cite{lippmann97}
which quotes a WER of 4\%, the 5.9\% estimate seemed rather high
(albeit measured on different data). This intriguing
discrepancy prompted us to launch our own human transcription effort
in order to confirm (or disconfirm) the estimates from~\cite{xiong16}.
The findings from this effort were doubly surprising.  First, we were
expecting the SWB measurement to be closer to the Lippmann estimate of
4\% but could only get down to 5.1\% for the best transcriber after
quality checks. Second, the same transcriber achieved a surprisingly
low 6.8\% WER for CallHome (we were expecting a much higher number
based on the 11.3\% estimate).

For comparison, our latest ASR system achieves 5.5\%/10.3\% WER on
SWB/CH.  This means that ``human parity'' is attainable on this
particular Switchboard subset (although not achieved yet) but is a
distant dream for the CallHome task. What makes the Switchboard and
CallHome testsets so different one might ask?  The biggest problem
with the SWB testset is that 36 out of 40 test speakers appear in the
training data, some in as many as 8 different
conversations~\cite{fiscus00}, and our acoustic models are very good
at memorizing speech patterns seen during training. The second problem
is that the SWB and CH tasks differ in the style of conversational
speech: SWB consists of conversations between strangers while CH
consists of calls between friends and family members.  Speaking style
between strangers tends to be more formal whereas the CallHome style
is more casual making CH a harder task. The training data collected by
LDC under the Switchboard and Fisher protocols is almost entirely
Switchboard-like meaning that testing on CallHome is a mismatched
scenario for ASR systems. Since ASR systems are generally not robust
to mismatched training and testing conditions, it comes as no surprise
that the degradation in performance from SWB to CH for ASR systems is
larger than that of expert transcribers.

On the system side, we have simplified and improved our acoustic
models considerably and experimented with more sophisticated language
models such as LSTM and WaveNet LMs. Most of the AM improvement comes
from LSTMs that operate on multiple features or use a different
training criterion such as speaker-adversarial multi-task learning.
Additionally, replacing the VGG convolutional
nets~\cite{simonyan2014very} that we had in our last year's
system~\cite{saon16} with ResNets~\cite{he2016deep} turned out to be
beneficial for performance. On the LM side, adding LSTM word and
character-based LMs resulted in substantial accuracy gains.

The rest of the paper is organized as follows: in section~\ref{human} we talk about
the human transcription experiments; in section~\ref{system}
we describe a series of system improvements pertaining to both
acoustic and language modeling and in section~\ref{conclusion} we
summarize our findings.

\section{Human transcription experiments}
\label{human}

These experiments were carried out by IBM's preferred speech
transcription provider, Appen. The transcription protocol that was
agreed upon was to have three independent transcribers provide
transcripts which were quality checked by a fourth senior transcriber.
All four transcribers are native US English speakers and were selected
based on the quality of their work on past transcription projects. The
transcribers were familiarized with the LDC transcription guidelines
which cover hyphenations, spelled abbreviations, contractions, partial
words, non-speech sounds, etc.

The transcription time was estimated at 12-14 times real-time (xRT) for the first pass for
Transcribers 1-3 and an additional 1.7-2xRT for the second quality
checking pass (by Transcriber 4). Both passes involved listening to the
audio multiple times: around 3-4 times for the first pass and 1-2 times for
the second. After receiving the transcripts, the following filtering
rules were aplied:

\begin{itemize}
\item All non-speech markers were tagged as non-lexical items which are ignored during scoring. Examples of non-speech markers are: [laughter], [breathing], [noise], \{no speech\}, etc.
\item Other markers such as '...', '--', '(( ))' were eliminated prior to scoring. 
\item All partial words ending in '-' were marked as non-lexical items.
\item All punctuation marks such as '.', ',', '!' and '?' were eliminated prior to scoring.
\end{itemize}

In order to use NIST's scoring tool {\tt sclite}, we had to convert
the transcripts into CTM files which have time-marked word boundary
information. This was done by splitting the duration of the utterance
uniformly across the number of words.

In Table~\ref{human-wer} we show the error rates of the three
transcribers before and after quality checking by the fourth
transcriber as well as the human WER reported in~\cite{xiong16}. Unsurprisingly, there is some variation among
transcriber performance and the quality checking pass reduces the error rate
across all transcribers. 

\begin{table}[htpb!]
\begin{center}
\begin{tabular}{|l|c|c|} \hline
                    & WER SWB & WER CH      \\ \hline
Transcriber 1 raw & 6.1 & 8.7        \\
Transcriber 1 QC & 5.6 & 7.8 \\ \hline
Transcriber 2 raw & 5.3 & 6.9        \\
Transcriber 2 QC & {\bf 5.1} & {\bf 6.8} \\ \hline
Transcriber 3 raw & 5.7 & 8.0        \\
Transcriber 3 QC & 5.2 & 7.6 \\ \hline\hline
Human WER from~\cite{xiong16} & 5.9 & 11.3\\ \hline
\end{tabular}
\end{center}
\caption{\label{human-wer}
Word error rates on SWB and CH for human transcribers before and after quality checking contrasted with the human WER reported in~\cite{xiong16}.}
\end{table}

Additionally, in Tables~\ref{subs} and~\ref{del-ins}, we take a closer look
at the most frequent substitution, deletion and insertion errors for
our system output and the best human transcript after quality
checking.  While many of the errors look similar to those reported
in~\cite{xiong16}, there is a glaring discrepancy in the frequency of
top deletions for CallHome between our human transcript and theirs.
This suggests that the very different estimates for the human error
rate for CallHome (6.8\% versus 11.3\%) can be attributed to a much
lower deletion rate for our best human transcript.

\begin{table}[htpb!]
\centering
\resizebox{\columnwidth}{!}{
\begin{tabular}{|l|l|l|l|} \hline
\multicolumn{2}{|c|}{SWB} & \multicolumn{2}{c|}{CH}\\ \hline
ASR & Human                  & ASR & Human \\ \hline
11: and / in      & 16: (\%hes) / oh & 21: was / is    & 28: (\%hes) / oh \\
9: was / is       & 12: was / is     & 16: him / them    & 22: was / is \\
7: it / that      & 7: (i-) / \%hes  & 15: in / and    & 11: (\%hes) / \%bcack\\
6: (\%hes) / oh   & 5: (\%hes) / a   & 8: a / the    & 10: bentsy / benji \\
6: him / them     & 5: (\%hes) / hmm & 8: and / in    & 10: yeah / yep\\
6: too / to       & 5: (a-) / \%hes  & 8: is / was    &  9: a / the\\
5: (\%hes) / i    & 5: could / can   & 8: two / to    &  8: is / was\\
5: then / and     & 5: that / it     & 7: the / a    &  7: (\%hes) / a\\
4: (\%hes) / \%bcack & 4: \%bcack / oh & 7: too / to    &  7: the / a\\
4: (\%hes) / am   & 4: and / in      & 6: (\%hes) / a    &  7: well / oh\\ \hline
\end{tabular}}
\caption{\label{subs}Most frequent substitution errors for humans and ASR system on SWB and CH.}
\end{table}

\begin{table}[htpb!]
\centering
\resizebox{\columnwidth}{!}{
\begin{tabular}{|l|l|l|l||l|l|l|l|} \hline
\multicolumn{4}{|c||}{Deletions} & \multicolumn{4}{c|}{Insertions}\\ \hline
\multicolumn{2}{|c|}{SWB} & \multicolumn{2}{c||}{CH} & \multicolumn{2}{c|}{SWB} & \multicolumn{2}{c|}{CH}\\ \hline
ASR & Human   & ASR & Human & ASR & Human & ASR & Human\\ \hline
30: it    & 19: i    & 46: i    & 20: i  & 13:  i   & 16: is   & 23: a   & 17: is\\ 
20: i    & 17: it   &  46: it   & 18: and& 10:  a   & 14: \%hes& 14:  is   & 17: it\\
17: that    & 16: and  & 39: and    & 15: it & 7: and    & 12: i & 11: i   & 16: and\\
16: a    & 14: that &  32: is   & 15: the& 7: of   & 11: and    & 10: are    & 14: have\\
14: and    & 14: you  & 26: oh    & 14: is & 6: you    & 9: it  & 10: you    & 13: a\\
14: oh    & 12: is   &  25: a   & 13: not& 5: do    & 6: do     & 9: the    & 13: that\\ 
14: you    & 12: the  & 20: to    & 10: a  & 5: the    & 5: have& 8: have    & 12: i\\
12: \%bcack    & 11: a   & 19: that    & 10: in & 5: yeah    & 5: yeah& 8: that    & 11: \%hes\\
12: the    & 10: of   & 19: the    & 10: that& 4: air   & 5: you & 7: and    & 10: not\\
11: to    & 9: have  & 18: \%bcack    & 10: to & 4: in    & 4: are & 7: it    & 9: oh\\ \hline
\end{tabular}}
\caption{\label{del-ins}Most frequent deletion and insertion errors for humans and ASR system on SWB and CH.}
\end{table}
\section{System improvements}
\label{system}
In this section we discuss the training data and testsets that were
used as well as improvements in acoustic and language modeling. The
training set for our acoustic models consists of 262 hours of
Switchboard 1 audio with transcripts provided by Mississippi State
University, 1698 hours from the Fisher data collection and 15 hours of
CallHome audio. In order to allay fears that we may be overfitting to
the Hub5 2000 testsets by extensively testing on them, we have decided
to report results on a variety of testsets. Since the RT'02, RT'03,
RT'04 and DEV'04f testsets have not been used in more than a decade,
we are fairly confident that performance improvements on these
testsets are indicative of real progress.  Statistics about all the
testsets used in the experiments are given in Table~\ref{testsets}.

\begin{table}[htpb!]
\begin{center}
\begin{tabular}{|l|c|c|c|} \hline
Testset     & Duration & Nb. speakers & Nb. words      \\ \hline
Hub5'00 SWB & 2.1h & 40 & 21.4K       \\ \hline
Hub5'00 CH  & 1.6h & 40 & 21.6K \\ \hline
RT'02       & 6.4h &120 & 64.0K \\ \hline
RT'03       & 7.2h &144 & 76.0K \\ \hline
RT'04       & 3.4h &72  & 36.7K \\ \hline
DEV'04f     & 3.2h &72  & 37.8K \\ \hline
\end{tabular}
\end{center}
\caption{\label{testsets}
Testsets that are used to report experimental results.}
\end{table}

In~\cite{soltau14}, we have shown that convolutional and
non-convolutional AMs have comparable performance and good
complementarity. Hence, the strategy for our previous
systems~\cite{saon15,saon16} was to use a combination of recurrent and
convolutional nets. For example, in last year's system we used a score
fusion of three models which share the same decision tree: unfolded
RNNs with maxout activations, LSTMs and VGG nets. This year, in order
to simplify and enhance the overall architecture, we eliminated the
maxout RNN, we improved the LSTMs and we replaced the VGG nets with
residual nets (ResNets).

\subsection{LSTM acoustic models}
\begin{figure}[htpb]
 \begin{center}
   \includegraphics[scale=0.7]{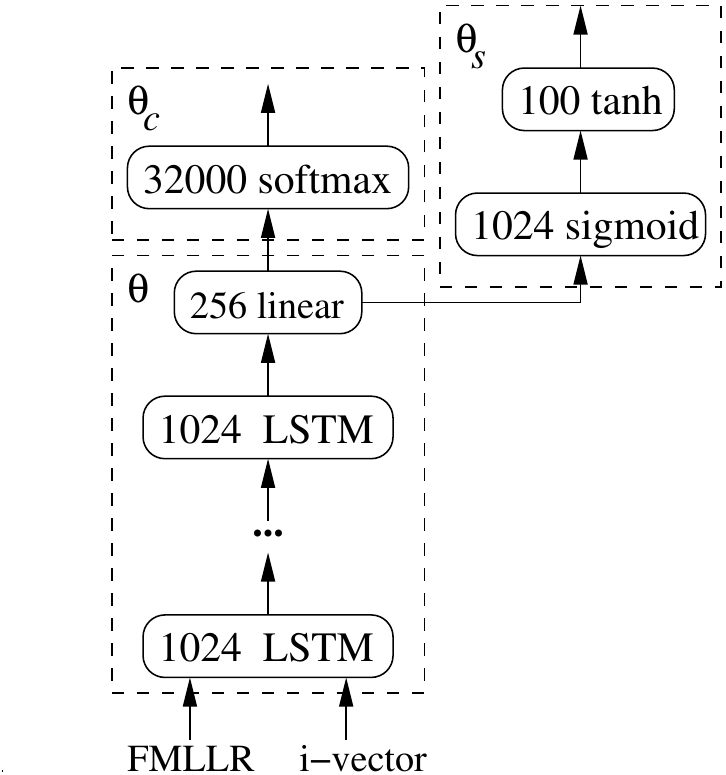}
       \caption{{\it LSTM with speaker-adversarial MTL architecture.}}
       \label{sa-mtl}
  \end{center}
\end{figure}

All models presented here share the following characteristics. Their
architecture consists in 4-6 bidirectional layers with 1024 cells per
layer (512 per direction), one linear bottleneck layer with 256 units
and an output layer with 32K units corresponding to as many
context-dependent HMM states (shown on the left side of
Figure~\ref{sa-mtl}). Training is done on non-overlapping subsequences
of 21 frames where each frame consists of 40-dimensional FMLLR
features to which we append 100-dimensional i-vectors. We group
subsequences from different utterances into minibatches of size 128
for processing speed and reliable gradient estimates. The training
consists of 14 passes of cross-entropy followed by 1 pass of SGD
sequence training using the boosted MMI criterion~\cite{povey08}
smoothed by adding the scaled gradient of the cross-entropy
loss~\cite{su2013error}. Implementation of the LSTM was done in Torch
\cite{collobert2011torch7} with cuDNN v5.0 backend.  Cross-entropy
training for each model took about 2 weeks for 700M samples/epoch, on a
single Nvidia K80 GPU device.

The first two improvements are fairly banal and consist in increasing
the number of layers from 4 (like in our previous model~\cite{saon16})
to 6 and in realigning the training data with a 6-layer LSTM and
retraining another LSTM. The effect of these steps is shown in the
first three rows of Table~\ref{lstm} across all testsets.

\begin{table}[htpb!]
\begin{center}
\resizebox{\columnwidth}{!}{
\begin{tabular}{|l|c|c|c|c|c|c|} \hline
LSTM                  & SWB  & CH   & RT'02 & RT'03 & RT'04 & DEV'04f\\ \hline
4-layer               & 8.0  & 14.3 & 12.2  & 11.6  & 11.0  & 10.8   \\ \hline
6-layer               & 7.7  & 14.0 & 11.8  & 11.4  & 10.8  & 10.4   \\ \hline
Realigned             & 7.7  & 13.8 & 11.7  & 11.2  & 10.8  & 10.2   \\ \hline
SA-MTL                & 7.6  & 13.6 & 11.5  & 11.0  & 10.7  & 10.1   \\ \hline
Feat. fusion          & 7.2  & 12.7 & 10.7  & 10.2  & 10.1  &  9.6   \\ \hline
\end{tabular}}
\end{center}
\caption{\label{lstm}
Word error rates for LSTM AMs across all testsets (36M n-gram LM).}
\end{table}

The second set of experiments was centered around the use of
speaker-adversarial multi-task learning (SA-MTL). In~\cite{ganin16},
the authors introduce domain-adversarial neural networks which are
models that are trained to not distiguish between in-domain, labeled
data and out-of-domain, unlabeled data. This is achieved by training a
domain classifier in parallel with the main classifier and by
subtracting the gradient component from the domain classifier when
estimating the parameters of the main classifier. This idea has been
successfully applied in speech by~\cite{shinohara16} in the context of noise
robustness where the author proposes noise-adversarial MTL to suppress
the effects of noise. Here, we experiment with training a speaker
classifier in addition to the main CD-HMM state classifier in order to
suppress the effects of speaker variability on ASR performance. Since
i-vectors are a good low-dimensional representation of a speaker, we
decided to train the speaker classifier to predict the i-vector inputs
using an MSE loss function. The speaker classifier has one sigmoid
layer and one hyperbolic tangent layer as shown in
Figure~\ref{sa-mtl}.

If we denote by $\theta$, $\theta_c$, $\theta_s$ the parameters of the common LSTM, the main classifier (weights of linear layer before softmax) and the speaker classifier, the SGD update is done according to:

\begin{eqnarray*}
 \hat{\theta}_c &=& \theta_c-\epsilon\frac{\partial{\cal
      L}_{CE}({\bf x})}{\partial\theta_c}\\
 \hat{\theta}_s &=& \theta_s-\epsilon\frac{\partial{\cal
      L}_{MSE}({\bf x})}{\partial\theta_s}\\
 \hat{\theta} &=& \theta-\epsilon\left(\frac{\partial{\cal
      L}_{CE}({\bf x})}{\partial\theta}-\lambda\frac{\partial{\cal
      L}_{MSE}({\bf x})}{\partial\theta}\right) 
\end{eqnarray*}
~~\\
where {\bf x} denotes a minibatch, ${\cal L}_{CE}$, ${\cal L}_{MSE}$ denote respectively the
cross-entropy loss of the main classifier and the mean-squared error
loss of the i-vector classifier, $\lambda$ is a scaling parameter
(typically set to 0.1), and $\epsilon$ is the learning rate. After the
model is trained, the i-vector classifier branch is discarded at test
time. As can be seen from Table~\ref{lstm} rows 3 and 4, we observe
some small gains across all testsets which are also due in part to
reestimating the VTLN warp factors and FMLLR transforms using an LSTM
decoding output (old factors and transforms were based off a GMM
decoding).

Last but not least, the largest improvement in LSTM modeling was
achieved through feature fusion. The thought process leading to this
experiment was that we wanted to add utterance-level information to
our models which were only looking at a window of 21 consecutive
frames. One possibility was to train an end-to-end LSTM using CTC as
in~\cite{miao15,sak15,soltau16,liu17} and append the features from the
last LSTM layer before the softmax to our existing features. This
experiment worked quite well however, upon closer inspection, it
turned out that the CTC model used a different set of input features:
Logmel+$\Delta$+$\Delta\Delta$ instead of PLP followed by LDA and
FMLLR. The question then naturally arose whether the gains came from
CTC modeling or from the different input representations. To answer
this question, we built an LSTM trained on fused
FMLLR+i-vector+Logmel+$\Delta$+$\Delta\Delta$ features the standard
way (without speaker-adversarial MTL). The WER improvement from adding the
Logmel features, indicated in Table~\ref{lstm} rows 3 and 5, is the
same as with CTC features meaning that the CTC modeling step was not
needed. Finally, we note that the feature fusion LSTM compares
favorably with other single acoustic models from the literature as
mentioned in~\cite{liu17} (Table 4).

\subsection{ResNet acoustic models}

On the convolutional network acoustic modeling side, we trained residual networks with pre-activation identity shortcut connections.
Residual Networks were introduced for image recognition in \cite{he2015deep} and used in speech recognition in \cite{xiong16, zhang2016very}.
The novelty of residual networks is to introduce shortcut connections between so-called ``blocks'' of convolutional layers, which was argued to improve the flow of information and gradients, and allows training even deeper CNNs without the optimization problem occuring without the residual connections.

\begin{table}[htpb!]
  \centering
\resizebox{\columnwidth}{!}{
  \begin{tabular}{| L{0.25\columnwidth} | R{0.30\columnwidth} | R{0.30\columnwidth} | R{0.30\columnwidth} | R{0.30\columnwidth}| }
\hline
        & (a)               & (b)           & (c)         & (d)    \\ \hline
Summary & Bottleneck 1-3333 & 1-3333 NoTimestride & 1-2222 Timestride & 1-3333 Timestride \\ \hline
\# param & 64.3 M & 67.1 M & 60.8 M & 67.1 M \\ 
\hline
Input & 
3 $\times$ 64 $\times$ 31 & 
3 $\times$ 64 $\times$ 55 & 
3 $\times$ 64 $\times$ 56 & 
3 $\times$ 64 $\times$ 76 \\
\hline
Stage~0 
64x32xT
& 
conv5x5,~64 
maxpool~(2x1)
&
conv5x5,~64 
maxpool~(2x1)
&
conv5x5,~64 
maxpool~(2x1)
&
conv5x5,~64 
maxpool~(2x1)
\\
\hline
Stage 1 
(64x32xT)
& 
\textit{initStride~1x1}
3x~[conv~1x1,~64 
conv~3x3,~64
conv~1x1,~256]
& 
\textit{initStride~1x1}
3x~[conv~3x3,~64
  conv~3x3,~64 ]
& 
\textit{initStride~1x1}
2x~[conv~3x3,~64
  conv~3x3,~64 ]
& 
\textit{initStride~1x1}
3x~[conv~3x3,~64
  conv~3x3,~64 ]
\\
\hline
Stage 2
(128x16xT)
& 
\textit{initStride~2x1}
3x~[conv~1x1,~128
conv~3x3,~128
conv~1x1,~512]
& 
\textit{initStride~2x1}
3x~[conv~3x3,~128
  conv~3x3,~128 ]
& 
\textit{initStride~2x1}
2x~[conv~3x3,~128
  conv~3x3,~128 ]
& 
\textit{initStride~2x1}
3x~[conv~3x3,~128
  conv~3x3,~128 ]
\\
\hline
Stage 3
(256x8xT)
& 
\textit{initStride~2x1}
3x~[conv~1x1,~256
conv~3x3,~256
conv~1x1,~1024]
& 
\textit{initStride~2x1}
3x~[conv~3x3,~256
  conv~3x3,~256 ]
& 
\textit{initStride~2x1}
2x~[conv~3x3,~256
  conv~3x3,~256 ]
& 
\textit{initStride~2x1}
3x~[conv~3x3,~256
  conv~3x3,~256 ]
\\
\hline
Stage 4
(512x4xT)
& 
\textit{initStride~2x1}
3x~[conv~1x1,~512
conv~3x3,~512
conv~1x1,~2048]
maxpool~(2x1)
& 
\textit{initStride~2x1}
3x~[conv~3x3,~512
  conv~3x3,~512 ]
maxpool~(2x1)
& 
\textit{initStride~\textbf{2x2}}
2x~[conv~3x3,~512
  conv~3x3,~512 ]
maxpool~(\textbf{2x2})
& 
\textit{initStride~\textbf{2x2}}
3x~[conv~3x3,~512
  conv~3x3,~512 ]
maxpool~(\textbf{2x2})
\\
\hline
Output 
& 
3x~FC~2084 \qquad\qquad
FC~1024\qquad\qquad
FC~32k
& 
3x~FC~2084 \qquad\qquad
FC~1024\qquad\qquad
FC~32k
& 
3x~FC~2084 \qquad\qquad
FC~1024\qquad\qquad
FC~32k
& 
3x~FC~2084 \qquad\qquad
FC~1024\qquad\qquad
FC~32k
\\
\hline
\hline
(XE-300) SWB & 11.8 & 11.2 & 11.3 &  11.4 \\ \hline
(XE) SWB     &      &  9.7 & 9.5  &  9.2  \\ \hline
(ST) SWB     &      & 8.6  & 8.7  & 8.3  \\ \hline
(ST) CH      &      & 15.5 & 15.0 & 14.9 \\ \hline
(ST) RT'02   &      & 13.4 & 13.3 & 13.1 \\ \hline
(ST) RT'03   &      & 13.1 & 12.7 & 12.7 \\ \hline
(ST) RT'04   &      & 12.1 & 12.0 & 11.9 \\ \hline
(ST) DEV'04f &      & 11.3 & 11.1 & 11.2 \\ 
\hline
\end{tabular}}
    \caption{ResNet architectures and results.
        Decoding with small LM (4M n-grams).
        In the bottom rows (results on test-sets). XE-300 indicates the network was cross-entropy trained on the 300h SWB corpus only,
        XE and ST for training on the 2000h SWB+Fisher corpus.
        Column (d) has best performance, compared against 3 different ablation variants: (a) with bottleneck blocks and without pooling, (b) without pooling, and (c) less depth.
        The size of the output of the $3 \times 3$ convolutions is indicated for each stage.
      }
      \label{tab:resnet} 
\end{table}

Table~\ref{tab:resnet} shows four residual network model architectures and their performance on the testsets with small LM.
We achieved best results with basic residual blocks without bottleneck, similar to the observations from \cite{zagoruyko2016wide} on CIFAR and SVHN experiments.
However, bottleneck residual blocks could possibly be optimal with a larger computational budget.
\nocite{he2016identity, sercu2015very, sercu2016advances, sercu2016dense} 
The input to our network are vtln-warped logmel features with 64 mel bins.
We perform data-balancing according to \cite{sercu2015very} with exponent $\gamma=0.8$.
We use full pre-activation identity shortcut connections which keep a clean information path \cite{he2016identity} without nonlinearity after addition.
For batch normalization the statistics are accumulated per feature map and per frequency bin following \cite{sercu2016dense}.

\begin{figure}[tb]
  \begin{center}
       \includegraphics[scale=0.35]{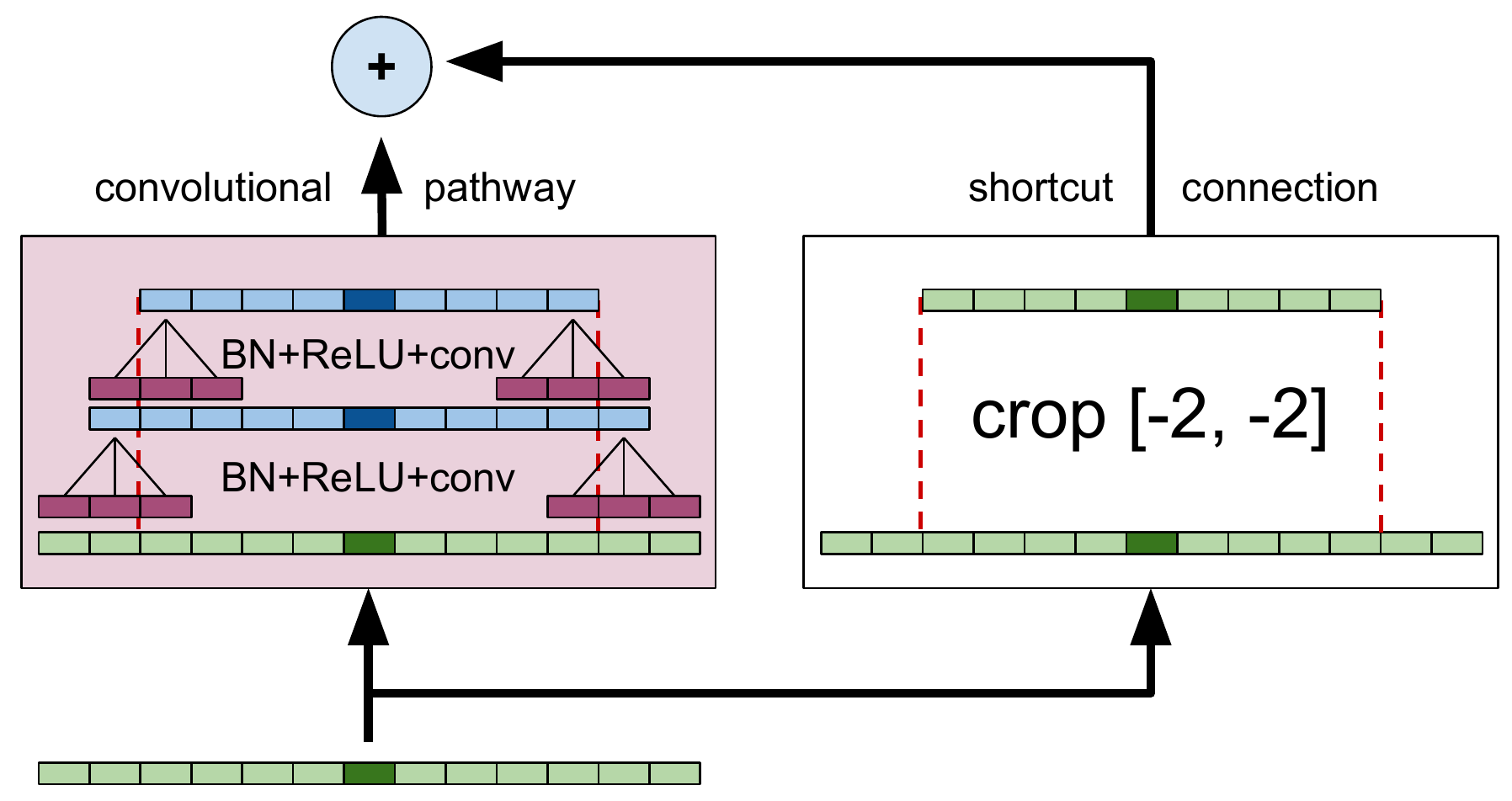}
       \caption{Residual connections on sequences. The convolutions are unpadded and reduce the size of the feature maps in the time direction (indicated with red dashed lines). 
       To match this reduction, we simply crop the edges along the time on the shortcut connection.}
       \label{fig:resnet}
  \end{center}
\end{figure}

In order to use residual networks for acoustic modeling, we need to adapt the residual blocks (see Figure \ref{fig:resnet}), while taking efficient convolution on sequences into account.
In ResNets for image classification the convolutional pathway only includes padded convolutions, so does not reduce the size of the feature maps.
The addition with the shortcut pathway is trivial, since both feature maps have the same size.
In contrast, for convolutions on sequences we can not pad along the time directions.
Padding along the time direction would modify the values on the edges based on the input sliding window location, thus making efficient convolution over a full utterance impossible (see \cite{sercu2016advances}).
So we do not pad the convolutions in time and as a consequence, the convolutional pathway reduces the size of the feature maps along the time direction.
In this case, we need to crop on the shortcut connection to match the size of the feature maps coming out of the convolutional pathway.
It is important to note that this does not impact the ability to convolve the residual net over full utterances at once:
since the values at the edges are computed the same as everywhere else, they are independent of the position of the input window.

Let us now consider how to use strided pooling and strided convolutions, and the relation to time-dilated convolutions.
First off, in the frequency direction, similar as for images, convolutions are padded so they do not reduce the size.
Rather, the size is reduced by a factor of 2 through convolutions with stride 2.
In Table~\ref{tab:resnet}, the ``initStride'' field on the first line of each stage indicates the (frequency x time) stride for the first block of that stage, where the number of feature maps is increased.
This stride applies to both the first $3 \times 3$ convolution of the block, and the $1 \times 1$ convolution in the projection shortcut.
The output feature map size is indicated in the left column for each stage.
\\
Secondly, along the time direction, strided convolutions and strided pooling is optional, but was found to improve performance \cite{sercu2016dense}.
In Table~\ref{tab:resnet}, Stage 4, column (c) and (d), bold indicates striding in time.
Note that, when adding time-strided conv and pool to an architecture, we need to increase the input context window to compensate for the additional size reduction.
For residual networks, similar as for VGG-style networks, we indeed observe that time-strided time-pooling improves performance, see column (b) vs (d).

When transitioning from cross-entropy (XE) to sequence training (ST), we want to modify our network to do dense prediction efficiently \cite{sercu2016dense}.
This means the intermediate states of the convolutional layers and output of the ResNet should maintain inputs full time-resolution, i.e. it should produce an output CD~state distribution for each input frame.
We can achieve this by using time-dilated convolutions according to the same recipe as in \cite{sercu2016dense}:
for each layer which originally strides in time with factor 2, set time-stride to 1 and dilate with factor 2 all consecutive convolutions, maxpooling and fully connected layers.
This includes the projection shortcut in the first block of each stage, though dilation for these $1 \times 1$ convolutions is irrelevant.
After these modifications, the residual net can be used for dense prediction on sequences.

The ResNet which we will use in further sections is in Table~\ref{tab:resnet} (d).
It has 12 residual blocks, 30 weight layers and 67.1~M parameters.
We trained this model using Nesterov accelerated gradient with learningrate 0.03 and momentum 0.99.
Implementation of the CNN was also done in Torch with cuDNN v5.0 backend.
Cross-entropy training took about 80 days for 1.5 billion samples, on 2 Nvidia K80 GPU's (4 devices) with batch size 64 per GPU and full synchronization between every minibatch.
We sequence trained this model for 200M frames with the boosted MMI criterion \cite{povey08}.

\subsection{Model combination}
In Table~\ref{comb} we report the performance of the best individual
models described in the previous paragraphs as well as the results
after frame-level score fusion across all testsets. All decodings are
done with an 85K word vocabulary and a 4-gram language model with 36M
n-grams. We note that LSTMs and ResNets exhibit a strong
complementarity which improves the WER for all testsets.

\begin{table}[htpb!]
\begin{center}
\resizebox{\columnwidth}{!}{
\begin{tabular}{|l|c|c|c|c|c|c|} \hline
Model              & SWB  & CH   & RT'02 & RT'03 & RT'04 & DEV'04f\\ \hline
LSTM1 (SA-MTL)     & 7.6  & 13.6 & 11.5  & 11.0  & 10.7  & 10.1   \\ \hline
LSTM2 (Feat. fusion) & 7.2  & 12.7 & 10.7  & 10.2  & 10.1  &  9.6   \\ \hline
ResNet             & 7.6  & 14.5 & 12.2  & 12.2  & 11.5  & 11.1   \\ \hline
ResNet+LSTM2       & 6.8  & 12.2 & 10.2  & 10.0  &  9.7  &  9.4   \\ \hline
ResNet+LSTM1+LSTM2 & 6.7  & 12.1 & 10.1  & 10.0  &  9.7  &  9.2   \\ \hline
\end{tabular}}
\end{center}
\caption{\label{comb}
Word error rates for LSTMs and ResNet and frame-level score fusion results across all testsets (36M n-gram LM).}
\end{table}

\begin{figure*}[tb]
  \begin{center}

   \begin{tabular}{c}

    \begin{minipage}{0.25\hsize}
      \begin{center}
       \includegraphics[scale=0.35]{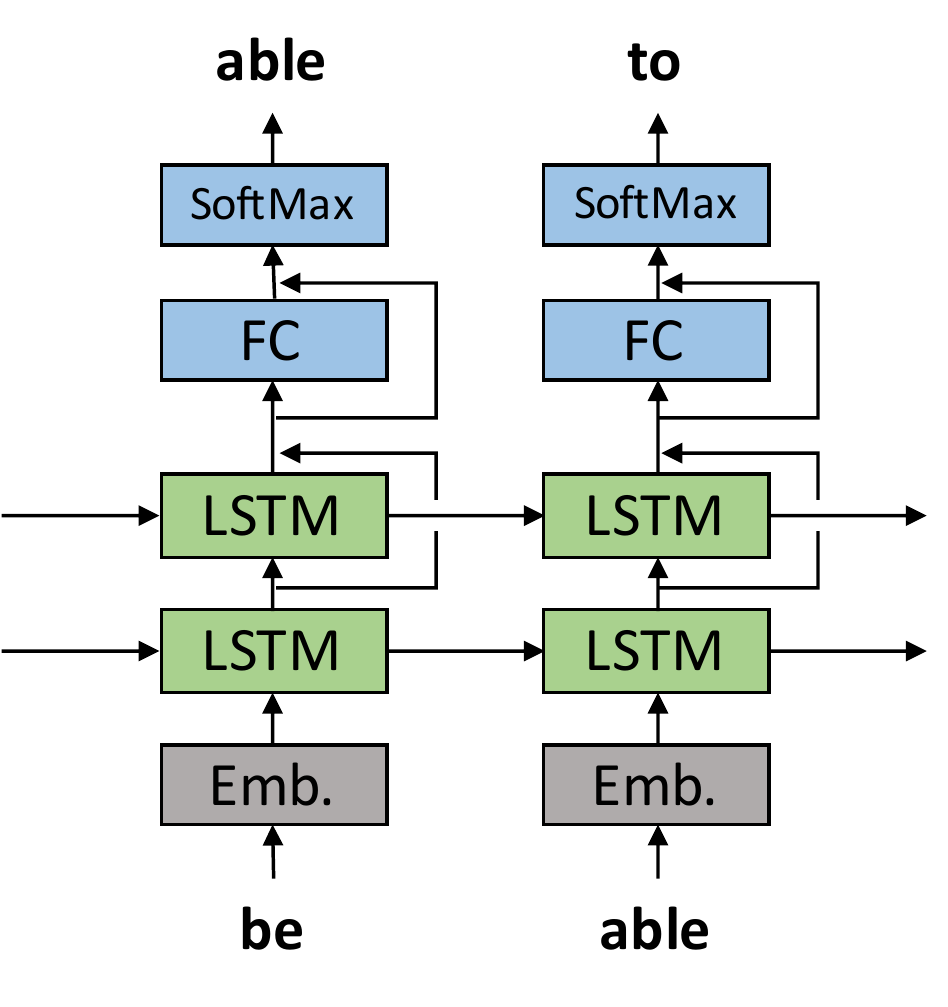}
       \caption{{\it Word-LSTM}}
       \label{fig:word_lstm}
      \end{center}
    \end{minipage}

    \begin{minipage}{0.0\hsize}
     ~~~~~
    \end{minipage}

    \begin{minipage}{0.4\hsize}
      \begin{center}
       \includegraphics[scale=0.35]{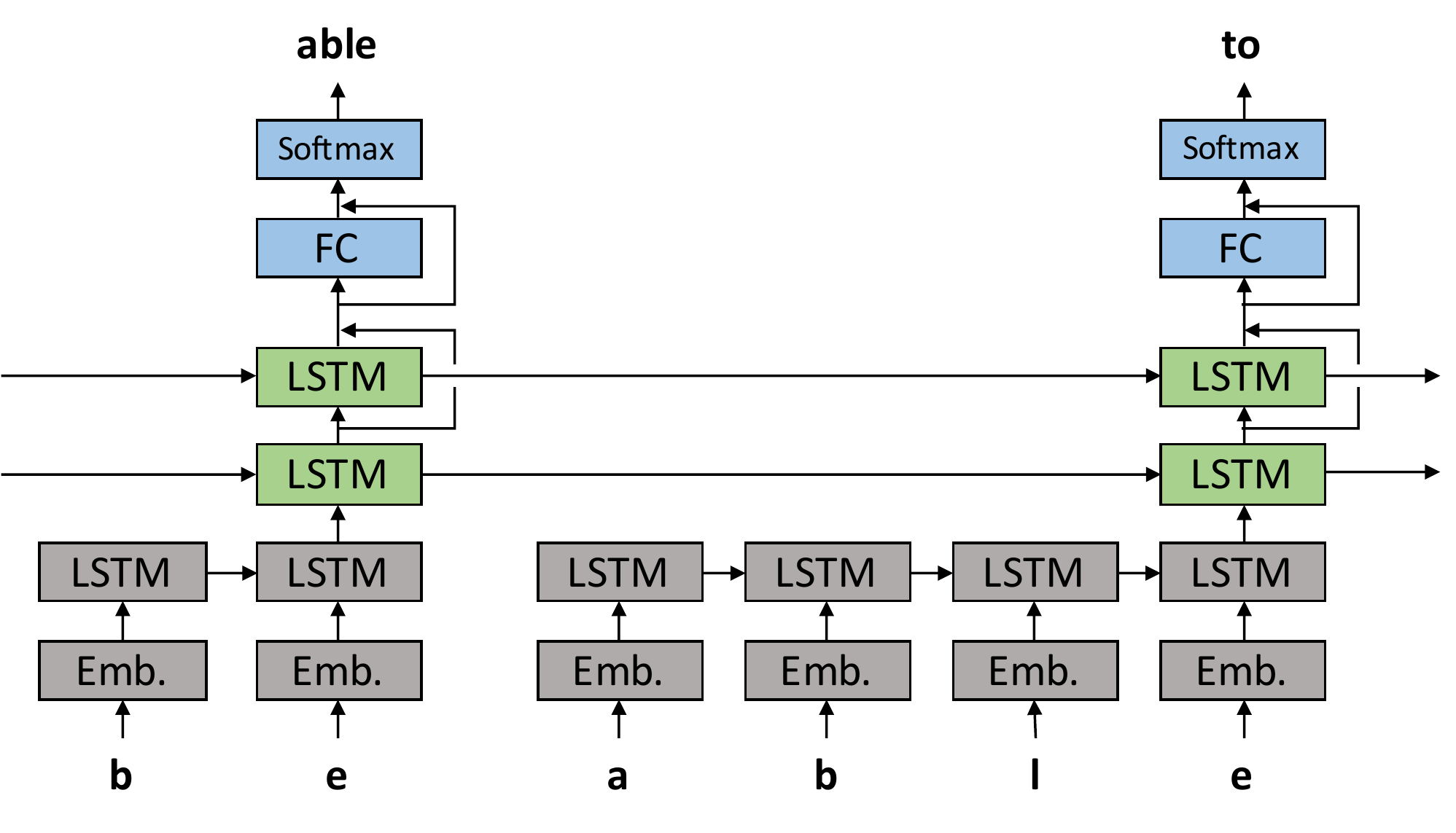}
       \caption{{\it Char-LSTM}}
       \label{fig:char_lstm}
      \end{center}
    \end{minipage}

    \begin{minipage}{0.0\hsize}
     ~~~~~
    \end{minipage}

    \begin{minipage}{0.35\hsize}
     \begin{center}
      \includegraphics[scale=0.35]{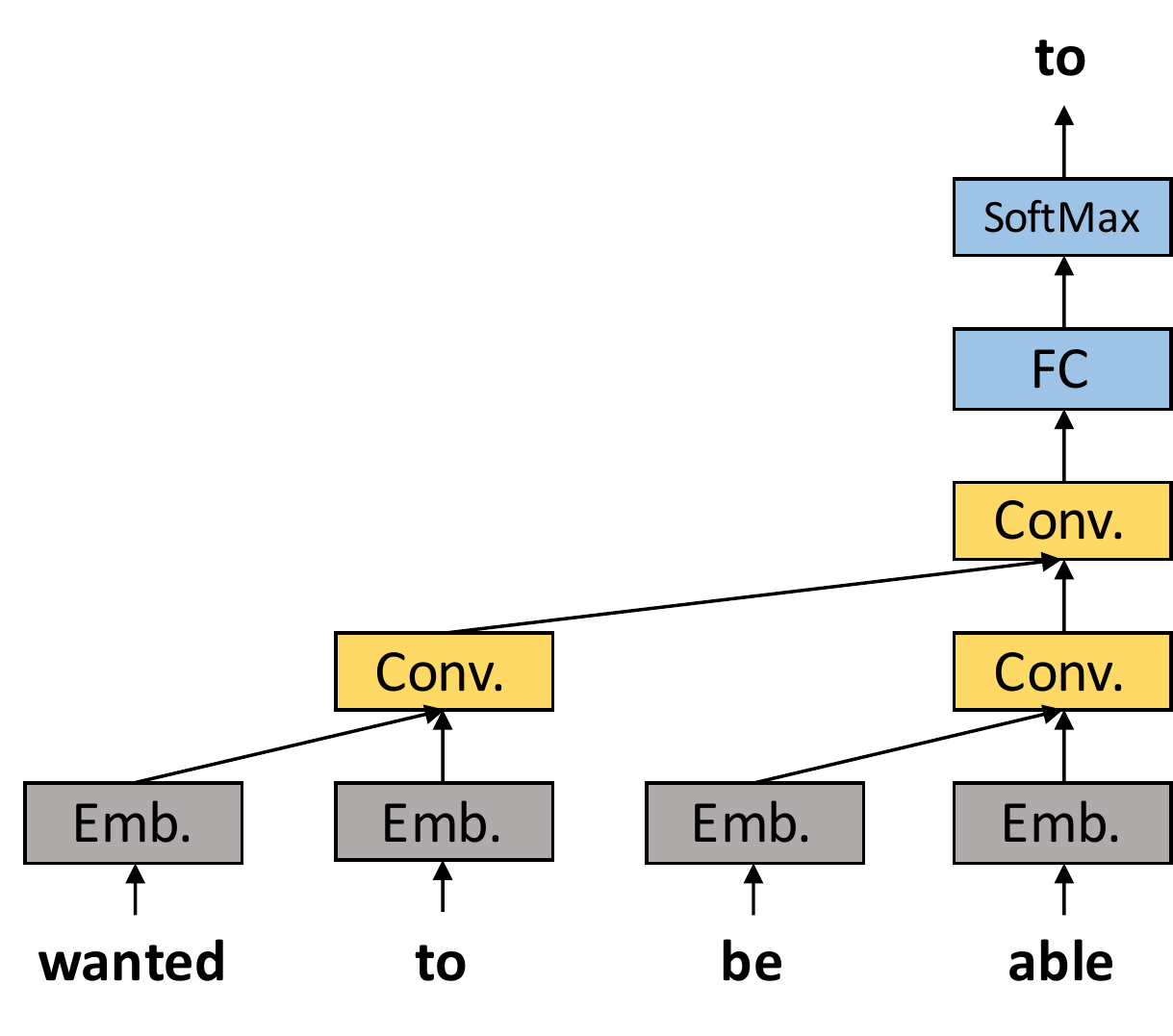}
      \caption{{\it Word-DCC}}
      \label{fig:word-dcc}
     \end{center}
    \end{minipage}
    
   \end{tabular}
   
  \end{center}
\end{figure*}
\begin{table}[t]
 \begin{center}
  \begin{tabular*}{\columnwidth}{@{\extracolsep{\fill}}lcc}
   \thline
   ~&\multicolumn{2}{c}{WER~[\%]} \\
   ~& SWB & CH \\
   \thline
   n-gram & 6.7 & 12.1 \\
   n-gram + model-M & 6.1 & 11.2 \\
   \hline
   n-gram + model-M + Word-LSTM& 5.6 & 10.4 \\
   n-gram + model-M + Char-LSTM& 5.7 & 10.6 \\
   n-gram + model-M + Word-LSTM-MTL& 5.6 & 10.3 \\
   n-gram + model-M + Char-LSTM-MTL& 5.6 & 10.4 \\
   n-gram + model-M + Word-DCC& 5.8 & 10.8 \\
   \hline
      n-gram + model-M + 4 LSTMs + DCC& {\bf 5.5} & {\bf 10.3} \\
   \thline
  \end{tabular*}
  \caption{\it WER on SWB and CH with various LM configurations.}
  \label{tab:lm}
 \end{center}
	\end{table}

\subsection{Language modeling improvements}
\label{lm}
In addition to n-gram and model-M used in our previous system~\cite{saon16}, we introduced LSTM-based as well as convolution-based LMs in this paper.

We experimented with four LSTM LMs, namely {\it Word-LSTM}, {\it
  Char-LSTM}, {\it Word-LSTM-MTL}, and {\it Char-LSTM-MTL}.  The
Word-LSTM had one word-embeddings layer, two LSTM layers, one
fully-connected layer, and one softmax layer, as shown in
Figure~\ref{fig:word_lstm}. The upper LSTM layer and the
fully-connected layer were wrapped by residual
connections~\cite{he2016deep}.  Dropout was only applied to the
vertical dimension and not applied to the time
dimension~\cite{zaremba2014recurrent}.  The Char-LSTM added an additional
LSTM layer to estimate word-embeddings from character sequences as
illustrated in Figure~\ref{fig:char_lstm}~\cite{ling2015finding}.
Both Word-LSTM and Char-LSTM used the cross-entropy loss of predicting the next word
given its history as objective function, similar to conventional LMs.
In addition, we introduced multi-task learning (MTL) in Word-LSTM-MTL
and Char-LSTM-MTL.  We first clustered the vocabulary using Brown
clustering~\cite{brown1992class}.  When training Word-LSTM-MTL and
Char-LSTM-MTL, weighted summation of cross-entropy of predicting next
word given its history and next class given its history was used as
objective function.

Inspired by the complementarity of convolutional and non-convolutional
acoustic models, we experimented with a convolution-based LM in the
form of dilated causal convolution as used in {\sc
  WaveNet}~\cite{oord2016wavenet}. The resulting model is called {\it
  Word-DCC} and consists of word-embeddings layer, causal convolution
layers with dilation, convolution layers, fully-connected layers,
softmax layer, and residual connections.  The actual number of layers
and dilation/window sizes were determined using heldout data
(Figure~\ref{fig:word-dcc} has a simple configuration for
illustration purposes).

For these five LMs, the training data and training procedures are common and described below:
\begin{itemize}
 \item We used the same vocabulary of 85K words from~\cite{saon16}.
 \item We first train the LM with a corpus of 560M words consisting of publicly available text data from LDC, including Switchboard, Fisher, Gigaword, and Brodcast News and Conversations. Then, starting from the trained model, we further train the LM with only the transcripts of the 1975 hours audio data used to train the acoustic model, consisting of 24M words.
 \item We controlled the leaning rate by ADAM~\cite{kingma2014adam} and introduced a self-stabilization term to coordinate the layer-wise learning rates~\cite{ghahremani16:_self}.
 \item For all models, we tuned the hyper-parameters based on the perplexity of the heldout data which is a subset of the acoustic transcripts. The approximate number of parameters for each model was 90M to 130M.
\end{itemize}

We first generated word lattices with the n-gram LM and our best acoustic model consisting of ResNet and two LSTMs.
Then we rescored the word lattices with the model-M and generated n-best lists from the rescored lattice.
Finally, we applied the four LSTM-based LMs and the convolution-based LM.
Note that LM probabilities were linearly interpolated and the interpolation weights of all LMs were estimated using the heldout data.

Table~\ref{tab:lm} shows WER on SWB and CH with various LM
configurations.  The LSTM-based LMs show significant improvements over
the strong n-gram + model-M results.  The Word-DCC also has a marginal
improvement over the n-gram + model-M.  The effect of multi-task
learning was confirmed especially on CH.  Among the five LSTM-based
and convolution-based LMs, word-LSTM-MTL achieved the best WER of
5.6\% and 10.3\% on SWB and CH respectively. By combining five LMs on
top of n-gram + model-M, we achieved 5.5\% and 10.3\% WER for SWB and
CH respectively. Lastly, we summarize the improvements due to the
various language model rescoring steps across all testsets in
Table~\ref{tab:lm2}. We noticed that the testset references have
inconsistent transcription conventions with regards to spellings which
are not followed by periods for SWB and CH (e.g. T V) and followed by
periods for the other testsets (such as T. V.). The last line of
Table~\ref{tab:lm2} shows the WERs when periods are removed from both
the references and system outputs by adding the filtering rules A. $=>$
A ... Z. $=>$ Z to the GLM file.

More details about the language modeling are
given in a companion paper~\cite{kurata17}.

\begin{table}[htpb!]
\begin{center}
\resizebox{\columnwidth}{!}{
\begin{tabular}{|l|c|c|c|c|c|c|} \hline
                   & SWB  & CH   & RT'02 & RT'03 & RT'04 & DEV'04f\\ \hline
n-gram             & 6.7  & 12.1 & 10.1  & 10.0  & 9.7   & 9.2   \\ \hline
+ model-M          & 6.1  & 11.2 &  9.4  & 9.4   & 9.0   & 8.8    \\ \hline
+ LSTM + DCC       & 5.5  & 10.3 &  8.3  & 8.3   & 8.0   & 8.0    \\ \hline
'.' removal        & 5.5  & 10.3 &  8.3  & 8.0   & 7.7   & 7.1   \\ \hline
\end{tabular}}
\end{center}
\caption{\label{tab:lm2}
Word error rates for the different LM rescoring steps across all testsets. Last line shows WERs after '.' removal from the references and system outputs.}
\end{table}


\section{Conclusion}
\label{conclusion}
We have presented a set of acoustic and language modeling improvements
to our English Switchboard system that resulted in a new record word
error rate on this task. On the acoustic side, two things were
instrumental in reaching this level of performance. The first one is a
steady improvement in bidirectional LSTM modeling, chief among them
being a simple feature fusion experiment. The second one is the
replacement of VGG nets with residual nets which are a more effective
architecture on the ImageNet classification task. When combined
together, these recurrent and convolutional nets show good
complementarity and enhanced accuracy on a variety of testsets. On the
language modeling side, we exploited the same complementarity between
recurrent and convolutional architectures by adding word and
character-based LSTM LMs and a convolutional WaveNet LM.

The second and perhaps more important point made in this paper is
that, unlike what was claimed in~\cite{xiong16}, we do not believe
that human parity has been reached on this task. The reasons why we
came to the opposite conclusion are twofold.  First, the Hub5'00 SWB
testset has a large overlap between training and test speakers which
results in ASR systems having deceptively good performance. A more
realistic level of ASR performance is the average WER across all
testsets which is around 8\% for our system. The second and more
direct argument is that the human WER of expert transcribers that were
asked to do a high-quality job is simply lower than what was
previously reported. On an optimistic note, this means that the future
of research in conversational speech recognition looks bright for at
least a few more years.

\bibliographystyle{IEEEtran}
\bibliography{inter2017}
\end{document}